\documentclass[a4paper]{article}

\usepackage{INTERSPEECH2021}

\usepackage{diagbox}
\usepackage{tabularx}
\usepackage{caption} 
\usepackage{subcaption} 
\usepackage[letterspace=-25]{microtype}
\usepackage{amsmath,graphicx}
\usepackage{makecell}
\usepackage{xcolor}
\usepackage{url}
\usepackage{multirow}
\usepackage{subcaption}
\usepackage[T1]{fontenc}
\usepackage{booktabs}
\usepackage{cite}

\newcommand{\newpara}[1]{\vspace{6pt}\noindent\textbf{#1}}

\usepackage{hyperref}

\hypersetup{
    colorlinks=true,
    linkcolor=purple,
    filecolor=magenta,      
    urlcolor=purple,
}

\title{Look Who's Talking: Active Speaker Detection in the Wild}

\name{\parbox{\linewidth}{\centering You Jin Kim, Hee-Soo Heo, Soyeon Choe, Soo-Whan Chung$^{*}$\thanks{$^{*}$The work was done while the authors were at Yonsei University.}, Yoohwan Kwon$^{*}$, \\ Bong-Jin Lee, Youngki Kwon, Joon Son Chung}}

\address{
  Naver Corporation, South Korea}
\email{youjin.kim117@navercorp.com}

\begin{document}

\maketitle
\begin{abstract}
In this work, we present a novel audio-visual dataset for active speaker detection in the wild. A speaker is considered \textbf{active} when his or her face is visible and the voice is audible simultaneously. Although active speaker detection is a crucial pre-processing step for many audio-visual tasks, there is no existing dataset of natural human speech to evaluate the performance of active speaker detection. 
We therefore curate the {\em Active Speakers in the Wild} (ASW) dataset which contains videos and co-occurring speech segments with dense speech activity labels.
Videos and timestamps of audible segments are parsed and adopted from VoxConverse, an existing speaker diarisation dataset that consists of videos in the wild. 
Face tracks are extracted from the videos and active segments are annotated based on the timestamps of VoxConverse in a semi-automatic way. 
Two reference systems, a self-supervised system and a fully supervised one, are evaluated on the dataset to provide the baseline performances of ASW.
Cross-domain evaluation is conducted in order to show the negative effect of dubbed videos in the training data.

\end{abstract}

\section{Introduction}
\label{sec:intro}

How can an AI-enabled agent interact with the user in a public area? 
There are many individuals in the crowd who can be confused as the user, and the agent must recognise who is talking to them in order to address the command.

\begin{figure}[ht!]
    \centering
    \includegraphics[width=0.925\linewidth]{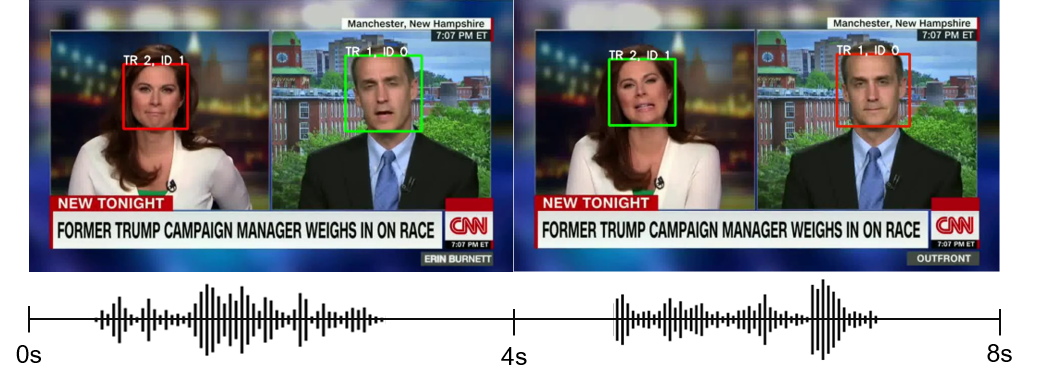}
    \caption{Who is speaking? ASD detects when a person is speaking. Active speakers are in green boxes, and inactive speakers are in red. Between $0$ and $4$ seconds, the woman on the left is not speaking (\textbf{active}), and the man on the right is  speaking (\textbf{non-active}). On the other hand, between $4$ and $8$ seconds, the woman on the left is speaking (\textbf{non-active}), and the man on the right is not speaking (\textbf{active}).}
\label{fig:asd}
\vspace{-10pt}
\end{figure}

Perceiving the rich information in a conversation is a core task in human computer interaction (HCI)~\cite{yan2016learning, gamboa2003identity, gamboa2004behavioral}, speech recognition~\cite{chung2020seeing, nassif2019speech, xiong2018microsoft, audhkhasi2017direct, noda2015audio}, speaker recognition~\cite{snyder2019speaker, xie2019utterance, snyder2018x, hoover2017putting} and video understanding~\cite{lin2019tsm, feichtenhofer2019slowfast, carreira2017quo}. It conveys identity, intents or emotions as well as the truth or knowledge. 
Active speaker detection (ASD) detects when a person speaks in a video, 
providing the crucial information on who is speaking to understand conversations in context (Figure~\ref{fig:asd}).

In addition to being an interesting problem in its own right, ASD is also an important pre-processing step for multi-modal speech processing -- it is a key component of pipelines to create VoxCeleb~\cite{nagrani2020voxceleb, chung2018voxceleb2, nagrani2017voxceleb} and LRS datasets~\cite{chung2016lip, Chung17, Chung17a}, and is essential for applications such as audio-visual speech enhancement~\cite{Afouras18, gabbay2018visual, hou2018audio, girin2001audio, chung2020facefilter, lee2021looking} and audio-visual speaker diarisation~\cite{ding2020self, chung2019said, gebru2017audio, friedland2009multi}.

ASD has some challenges in both audio and visual elements. Audio can contain background noises such as laughter, clapping, mumbling, camera shutter sound, and overlapping speech. Visual clues can be loss due to the following reasons. A speaker can turn his/her head aside or down while talking and cover the face by his/her hand. Also, some words are pronounced by only tongue movements. Therefore, instead of utilizing audio and visual elements individually, both information need to be used at the same time.

AVA-ActiveSpeaker~\cite{roth2020ava} is a recently released audio-visual dataset for ASD. It contains large amount of conversational videos, but is not suitable for the aforementioned tasks. 
A key limitation of this dataset is that a large proportion of the videos are dubbed movies, often into another language. Therefore, the audio and the video are not in correspondence or synchronisation. Without correspondence between audio and video, the network learns to detect whether there is speech in the audio and whether the lips are moving, but not whether the two correspond ({\em i.e.} originate from the same person). Although the sections of voice-over or dubbing are labelled as `positive' in the AVA-ActiveSpeaker, in the context of audio-visual speech enhancement, speaker recognition and many related applications, they should be considered negatives. For example, a model trained on this dataset would predict a silent video of Donald Trump's interview with the voice-over of a CNN narrator as positive.
These issues in the existing dataset raise the necessity for a new dataset of natural human speech.

In this paper, we present {\em Active Speakers in the Wild} (ASW), an audio-visual dataset for ASD that overcomes the limitations in the previous work.  
The dataset is created using a semi-automatic pipeline and will be released to the public. 

The videos in this dataset are based on 
VoxConverse~\cite{chung2020spot}, an audio-only dataset for speaker diarisation, which solves the problem of ``\textit{who spoke when}''.
The same list of video is used to create ASW, in order to make use of the speech activity labels from the VoxConverse dataset. 

In order to perform ASD using audio and video at the same time, a framework that can effectively model information common to both signals is required. 
The speech activity labels are combined with SyncNet~\cite{Chung16a} predictions to generate initial annotations, which are then corrected by human annotators. The annotations are double-checked in order to minimise human errors.

We provide a number of baselines for this task, including a SyncNet-based ASD and a pre-trained model from the winning entry for the ASD track of the ActivityNet challenge in 2019. 
We report a number of metrics including average precision (AP), area under the receiver operating characteristic (AUROC), and equal error rate (EER).

\section{Dataset Description}
\label{sec:pagestyle}

The ASW dataset consists of 212 videos randomly selected from the VoxConverse dataset. The videos are divided into three sets, 106 for development, 53 for validation and 53 for testing. Face bounding-boxes are detected for every frame of 212 videos and are grouped together into face tracks. As a result, $30.9$ hours are generated and annotated. 
The development set has $13.4$ hours in which $7.6$ hours ($56.7\%$) are active tracks, and $5.8$ hours ($43.3\%$) are non-active tracks.
The validation set consists of $9.6$ hours where $5.8$ hours ($60.4\%$) are active, and $3.8$ hours ($39.6\%$) are non-active.
The test set is comprised with $7.9$ hours where $4.5$ hours ($57.0\%$) are active, and $3.4$ hours ($43.0\%$) non-active.
The summary of the statistics can be found in Table~\ref{tab:stat}.

The total number of the face tracks is $11551$, and the number of the tracks in development, validation, and test set are $4676$, $3483$ and $3392$, respectively. 
The minimum, maximum, mean length of the tracks from the development set are $0.2$, $233.0$, $10.3$ seconds, the validation set are $0.8$, $310.2$, $9.8$ seconds, and the test set are $0.8$, $153.8$, $8.4$ seconds. The summary of the statistics can be found in Table~\ref{tab:stat2}. 

\begin{figure}[t]
  \centering
  \includegraphics[width=0.725\linewidth]{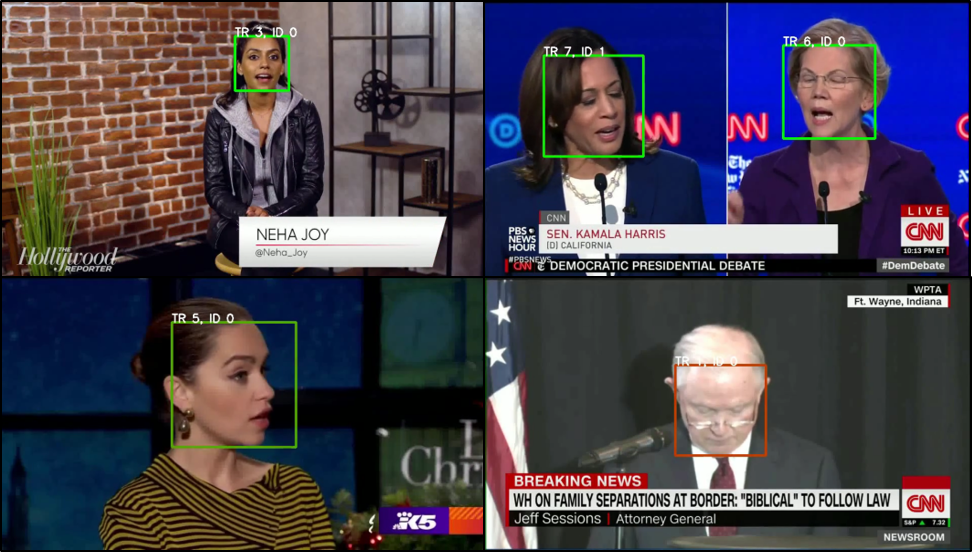}
  \caption{Examples from ASW.
  The videos are recorded in the wild, including political debates, talk-shows, press conferences and so on. It leads various types of audio challenges, such as laughter, applaud, cross-talking and visual challenges, such as profile faces and changes in light. (Green and red boxes indicate active or non-active.)}
  \label{fig:asw}
\end{figure}

Videos in the dataset are recorded in the wild and include news, political debates, press conferences, panel discussions, interviews, talk-shows and so on. Therefore, it contains various types of background noise, such as laughter, applause and camera shutter sound, as well as visual challenges such as occlusion and the deterioration of video quality as shown in Figure~\ref{fig:asw}.

Annotations are binary - \textbf{active} (positive) and \textbf{non-active} (negative). 
An audio-visual segment in a face track is active when it is audible and visible. 
The segment is considered to be audible when the utterance can be transcribed into words, and visible when a face detection appears in the face track. 
The ASW dataset will be released publicly, including the annotations.

\begin{table}[t]
  \caption{Statistics of ASW. The number of videos each set has are 106, 53, and 53 respectively. The duration of the active tracks and the non-active tracks are denoted in \textbf{hours}, and the ratio is in the parentheses.}
  \label{tab:perf_base}
  \centering
  \begin{tabular}{l|c|c|c}
    \toprule
    \textbf{Set} & \textbf{\# of videos} & \textbf{Active}  & \textbf{Non-active} \\
    \midrule
    Dev & 106 & 7.6 (56.7\%) & 5.8 (43.3\%) \\ \hline
    Val & 53 & 5.8 (60.4\%) & 3.8 (39.6\%) \\ \hline
    Test & 53 & 4.5 (57.0\%) & 3.4 (43.0\%) \\ 
    \bottomrule
  \end{tabular}
\label{tab:stat}  
\end{table}

\begin{table}[t]
  \caption{Statistics of face tracks from ASW. The number of tracks of each set is summarised. The min, max and mean of the duration of the track are denoted in \textbf{seconds}.}
  \label{tab:perf_base}
  \centering
  \begin{tabular}{l|c|c|c|c}
    \toprule
    \textbf{Set} & \textbf{\# of face tracks} & \textbf{Min}  & \textbf{Max} & \textbf{Mean}\\
    \midrule
    Dev & 4,676 & 0.2 & 233.0 & 10.3\\ \hline
    Val & 3,483 & 0.8 & 310.2 & 9.8\\ \hline
    Test & 3,392 & 0.8 & 153.8 & 8.4\\
    \bottomrule
  \end{tabular}
\label{tab:stat2}  
\end{table}

\section{Dataset Creation}
\label{sec:4}

\subsection{Automatic pipeline}
\label{ssec:4-1}

The automatic part of the pipeline generates initial annotations to be checked by human annotators.

\subsubsection{Face track extraction}
\label{sssec:4-1-1}

The visual pipeline to create the dataset is based on VoxConverse. The key stages are described in the following paragraphs.

\newpara{(1) Shot detection.} 
A video is split when the scene or camera angle changes. A shot boundary detection library, PySceneDetect~\cite{castellano2018pyscenedetect} is used, and it splits a video based on intensity or brightness changes. 

\newpara{(2) Face detection.} 
The face in a clip are spot by Single Shot Scale-invariant Face Detector (S3FD)~\cite{zhang2017s3fd}. It can detect small or turned faces and achieves high performance on several benchmark datasets. 

\newpara{(3) Face tracking.} 
The face detected by S3FD are concatenated, based on the position in consecutive frames. Intersection over union (IoU) defined as diving the area of overlap to the area of union between bounding-boxes is calculated between face bounding-boxes of the previous and the current frames. 
Pair of faces that has high IoU score are joined to form a track.

\subsubsection{Active speaker annotation}
\label{sssec:4-1-2}
 

We can find speech activity labels from the RTTM file of VoxConverse, since the RTTM files contain the start and the duration of all speaking segments.
Speech activity labels from the VoxConverse labels must be positive for any speaking segment, but there can be no corresponding speaking segment for some speech activity regions if the face of the speaker is not visible.

SyncNet has been proposed in~\cite{Chung16a} and has shown good performance in ASD.
SyncNet has originally been trained for audio-to-video synchronisation, the task of aligning the two modalities by locating the most relevant audio segment given a short video stream. 
For that purpose, this system learns audio and video signals as vectors in the joint embedding space, sharing the common linguistic information across two modalities.
It predicts whether or not the visible speaker is speaking based on the correlation between the audio and the video embeddings. 
The correlation is high when a speaker is visible in the audible segment and low in other cases. 
The segments of active speech are detected when the correspondence is larger than a pre-defined threshold, and the speech activity labels from the VoxConverse dataset is positive.

\subsection{Manual verification}
\label{ssec:4-2}
To minimise errors in the dataset, 6 human annotators have gone over the dataset to correct the errors from the automatic annotation and cross-check others' corrections. In all processes, the multimedia variation of VGG image annotator (VIA)~\cite{dutta2016vgg} is used.

\subsubsection{Guidelines and tool}
\label{sssec:4-2-1}

The result of the automatic detection is verified by human annotators.
The guideline for the audio stream is adopted from VoxConverse~\cite{chung2020spot}, while for the visual stream is newly defined. 
The speech is considered to be audible if it can be written down in words.
The video is defined as visible when a speaker appears, even if the speaker's mouth is barely visible. A speaker sometimes covers his/her lips with a hand, or turns his/her face side ways. It is still defined as visible, as long as the speaker's head appears in the video.  
A segment is labeled as \textbf{active}, when it is audible and visible at the same time. 

Original videos are provided, on which face bounding-boxes and prediction of SyncNet-ASD are overlaid.
Annotators are asked to annotate the true boundary of an active segment to $0.1$-second accuracy and divide a segment into two parts when there is a pause longer than 0.25 second.

\subsubsection{Cross-checking}
\label{sssec:4-2-2}

The human corrections are cross-checked by the other annotators.
Disagreements between the model and the annotators usually occur when the segment is particularly challenging. The second annotator is asked to re-check the videos, with particular attention to the sections modified by the first annotator.

\section{Active Speaker Detection}
\label{sec:majhead}
In this section, we describe two baseline methods for ASD, one using a cross-modal representation model trained by self-supervision and the other adopting a two-way classifier trained on an external dataset.

\subsection{Self-supervised method}
\begin{figure}[t]
  \centering
  \includegraphics[width=0.75\linewidth]{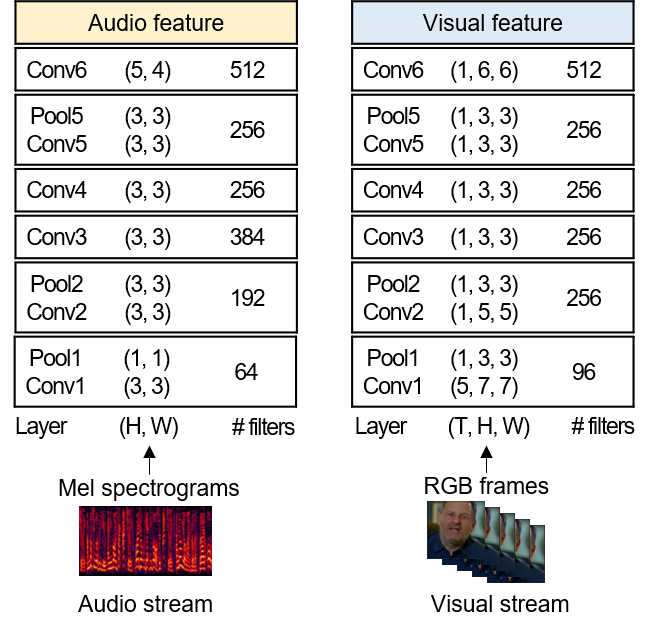}
  \caption{Architecture of the self-supervised method. It consists of audio and visual streams, trained using 0.2 second inputs. 40-dimensional mel-spectrogram is used as the input of an audio stream and RGB frames are used as the input of a visual stream.} 
  \label{fig:fe}
  \vspace{-8pt}
\end{figure}

In~\cite{Chung16a,Chung19}, the authors propose SyncNet to learn audio-visual representation using cross-modal self-supervision. The network configurations are shown in Figure~\ref{fig:fe}.
The audio and visual streams are trained in a multi-way matching method~\cite{Chung19,chung2020perfect} for the audio-to-video synchronisation task, where we use 40 candidates; 1 positive pair and 39 negative pairs.
The networks are trained to minimise the distance between positive embeddings and maximise the distance between negative embeddings simultaneously, using a multi-way matching (MWM) loss.
Its training criterion is as follows,
\begin{equation}
\label{eq:xent}
    \begin{gathered}
        L_{MWM}=-\frac{1}{2N}\sum_{n=1}^{N}\sum_{m=1}^{M}y_{n,m}\text{log}(p_{n,m})\\
        p_{n,m}=\frac{\exp(d_{n,m}^{-1})}{\sum_{k=1}^{M}\exp(d_{n,k}^{-1})},
    \end{gathered}
    \vspace{-3pt}
\end{equation}
where $d_{n,m}$ is the pairwise distance between audio and visual embeddings, and $y_{n,m}$ is a similarity metric where $1$ and $0$ indicate positive and negative pairs, respectively. 
$N$ is the number of samples, and $M$ is the number of candidates.  
In face tracks, active segments can be detected where the speech and the face of a person are synchronised, showing high similarity. 
We adopt either the inverse of L2 distance or cosine similarity to calculate the similarity score.

\subsection{Supervised method}
\cite{chung2019naver} uses SyncNet as a front-end feature extractor, and adds a back-end classifier to predict explicit labels. 
The back-end classifier is illustrated in Figure~\ref{fig:backend}. It consists of 2 bi-directional gated recurrent units (BGRUs)~\cite{cho2014learning} and 2 fully-connected (FC) layers. 
GRUs are adopted instead of long short-term memory (LSTM)~\cite{hochreiter1997long} used in~\cite{chung2019naver} for efficiency. 
The joint representation is derived by concatenating GRU outputs along the channel axis. 
It is then projected into a two node output layer for binary classification; label $1$ is for active, while label $0$ is for non-active frames.
Note that the cross-entropy loss is only used to train the back-end classifier, not to fine-tune the SyncNet layers.

\section{Experiments}
\label{sec:exp}

In this section, the performances of the reference systems on the ASW dataset are evaluated using both self-supervised and supervised methods.
We use three metrics to evaluate the ASD performance - AP, AUROC, and EER.

\newpara{Self-supervised method.} 
SyncNet extracts audio and visual features from $0.2$ second segment, moving $0.04$ second at a time. 
The similarities between audio-visual features are measured using either the inverse of L2 distance or cosine similarity.
An audio-visual pair is determined to be active, when the similarity score of it is higher than a pre-defined threshold.

\begin{figure}[t]
\vspace{-10pt}
  \centering
  \includegraphics[width=0.55\linewidth]{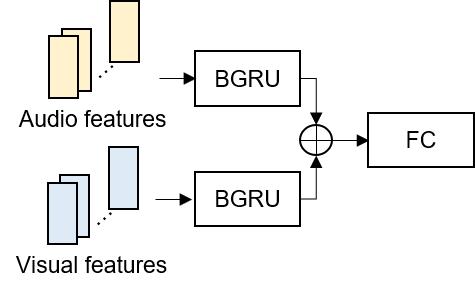}
  \caption{Back-end classifier of the supervised method. The audio and the visual features are generated from the front-end feature extractor, SyncNet, and ingested into BGRU layers. The outputs of BGRU are concatenated and fed into two FC layers, in order to determine the activeness.} 
  \label{fig:backend}
\end{figure}

\newpara{Supervised method.} 
Here, we adopt the front-end feature extractor from the self-supervised method and add the back-end classifier on the top of it. The back-end classifier consists of BGRUs and FC layers, as shown in Figure~\ref{fig:backend}. 
BGRUs have two layers with $128$ nodes each. 
FC layers have $128$ and $2$ nodes, respectively. 
The front-end feature extractor is fixed, and only the back-end classifier is fine-tuned. 
The numbers of active and non-active labels are balanced in a batch, training the back-end classifier. 

\newpara{Results.}
Table~\ref{tab:evalasw} shows the performance of two methods. 
The use of cosine similarity shows better performance than the inverse of L2 distance in case of the self-supervised method. 
The supervised method shows $5.5\%$ higher AP than the self-supervised method.

\begin{table}[t]
  \caption{Reference performances on the ASW dataset using the self-supervised and supervised methods. The similarity score of the self-supervised method is measured by the inverse of L2 distance or cosine similarity. The supervised method is a variant of the Activity challenge winning system in 2019 using BGRUs.}
  \label{tab:perf_base}
  \centering
  \begin{tabular}{l|l|c|c|c}
  \toprule
  \multicolumn{2}{l|}{\textbf{method}} & \textbf{AP} & \textbf{AUROC} & \textbf{EER} \\ 
  \midrule
\multirow{2}{*}{Self-supervised} & L2 &0.908  &0.955  &0.100 \\ \cline{2-5} 
                                 & cosine &0.924  &0.962  &0.083 \\ \hline
\multicolumn{2}{l|}{Supervised} &0.979  &0.988  &0.048 \\ 
\bottomrule
\end{tabular}
\label{tab:evalasw}  
\end{table}

\subsection{Discussion}
\label{sec:dis}

One of the main properties that differentiates the ASW dataset from the AVA-ActiveSpeaker dataset is that ASW does not include the dubbed videos, in which the audio is recorded on different languages.
Note that the dubbed videos in the AVA-ActiveSpeaker dataset are annotated as active frames without any indicator. 
When ASD is used as a pre-processing step for audio-visual speech processing, the positive labels allocated to dubbed videos can be regarded as mislabeled, and we hypothesise that it might cause the trained network to produce more false alarms. 
We propose two experiments to show the effect of the dubbed videos.

First, we conduct a cross-domain evaluation. 
Specifically, an identical model is trained using each dataset, and a threshold is tuned to meet EER using the corresponding validation set.
FAR and FRR of ASW are calculated using the test set to avoid overfitting, whereas FAR and FRR of AVA-ActiveSpeaker are calculated using the validation set, as the annotation of test set is not available. 
Then we observe how the FAR and FRR change when the evaluation sets are the same. 
Table~\ref{tab:cross} describes the results. 
Comparing the first and the fourth row, we can analyze the effect of different training datasets on ASW dataset. 
It is acceptable that the total error increases because of domain differences of datasets.
However, it is noteworthy that the specific type of error has increased significantly. 
In particular, in the model trained with the AVA-ActiveSpeaker dataset, the proportion of FAR has increased significantly, and we interpret that this result is caused by the dubbed videos in the AVA-ActiveSpeaker dataset. 
This is because, in the training process using the AVA-ActiveSpeaker dataset, dubbed videos with no correspondence between audio-visual pairs are forced to be classified as active frames, making the model easily mis-detect non-active frames.
We also find the same tendency by comparing the second row and the third row.

The second experiment uses five dubbed videos downloaded from YouTube. 
Note that these videos are included in neither ASW nor AVA-ActiveSpeaker.
In the viewpoint of speech pre-processing, all active frames detected from the dubbed video are false alarms because there is no correspondence between the video and the audio streams. 
Therefore, we expect that the precision of the method trained by ASW or AVA-ActiveSpeaker can be estimated using the ratio of detecting active frames from the dubbed video (\textit{lower is better}). 
Table~\ref{tab:case} shows the ratio of active frames detected by the method trained by each dataset and their average value. 
The method trained by ASW shows $19\%$ lower false alarms in average.
It shows that false positive labels included in AVA-ActiveSpeaker reduce the precision of the system.

\begin{table}[t]
  \caption{Cross-domain evaluation. The thresholds of the first and the third row are adopted to the second and the fourth row, respectively, in order to see how the FAR and FRR changes. (AVA: AVA-ActiveSpeaker)}
  \vspace{-3pt}
  \centering
  \begin{tabular}{l|l|c|c|c}
  \toprule
  \textbf{Train data} & \textbf{Eval data} & \textbf{Thres.} & \textbf{FAR} & \textbf{FRR} \\
  \midrule
  ASW & ASW & 0.60 & 0.04 & 0.07 \\ \hline
  ASW & AVA & 0.60 & 0.03 & 0.73 \\ \hline
  AVA & AVA & 0.43 & 0.18 & 0.18 \\ \hline
  AVA & ASW & 0.43 & 0.31 & 0.06 \\ 
  \bottomrule
  \end{tabular}
\label{tab:cross}  
\end{table}

\begin{table}[t]
  \caption{The ratio of active frames detected in dubbed videos. All frames in dubbed videos should not be detected as active, since there is no correspondence between audio and visual streams. (AVA: AVA-ActiveSpeaker) }
  \vspace{-3pt}
  \centering
  \begin{tabular}{l|c|c}
    \toprule
    \diagbox[width=10em]{\textbf{Video id}}{\textbf{Train data}} & \textbf{ASW} & \textbf{AVA} \\ 
    \midrule
    \tt{1odS6ynbWNo} & 0.20 & 0.31 \\ \hline
    \tt{KQViP7O8t98} & 0.40 & 0.67 \\ \hline
    \tt{krCpn6RrNX8} & 0.24 & 0.39 \\ \hline
    \tt{x3D6dr-feaU} & 0.26 & 0.48 \\ \hline
    \tt{ZiOJq6PMkls} & 0.12 & 0.28 \\ 
    \midrule
    \textbf{Average} & 0.24 & 0.43 \\ 
    \bottomrule
  \end{tabular}
\label{tab:case}  
\end{table}

\section{Conclusion}
\label{sec:con}

In this paper, we present a novel ASD dataset, ASW, and the corresponding reference systems. 
ASW contains $30.9$ hours of videos, in which positive and negative labels are balanced. 
The baseline performances are measured by AP, AUROC and EER, adopting the self-supervised and supervised systems. 
ASW does not include dubbed videos unlike the existing dataset, AVA-ActiveSpeaker.  
Cross-domain evaluation and case study show that the dubbed videos cause the threshold shift and lower the precision of the systems. 
We expect that ASW will be well-applicable to real-world scenarios since it consists of natural human speech and does not contain false positive labels.

\section{Acknowledgements}
We would like to thank Jee-weon Jung, Jung A Choi and Icksang Han for organising the annotations and for helpful discussions.

\clearpage
\bibliographystyle{IEEEtran}
\bibliography{longstrings,mybib,vgg_local}

\end{document}